\documentclass[runningheads]{llncs}
\usepackage{graphicx}

\usepackage{tikz}
\usepackage{comment}
\usepackage{amsmath,amssymb} 
\usepackage{color}
\usepackage{footnote}

\usepackage{graphicx}
\usepackage{booktabs}
\usepackage{soul}
\usepackage{color,soul}
\usepackage{enumitem} 
\usepackage{subcaption} 
\usepackage{bbm}
\usepackage{multirow}
\usepackage{array}
\usepackage{breakcites}

\usepackage{adjustbox}
\usepackage[pagebackref=true,breaklinks=true,letterpaper=true,colorlinks,bookmarks=false]{hyperref}
\usepackage{pdfpages}

\newcommand{\ie}{\textit{i}.\textit{e}.}
\newcommand{\eg}{\textit{e}.\textit{g}.}

\DeclareMathOperator*{\argmax}{argmax}

\newcolumntype{L}[1]{>{\raggedright\let\newline\\\arraybackslash\hspace{0pt}}m{#1}} 
\newcolumntype{R}[1]{>{\raggedleft\let\newline\\\arraybackslash\hspace{0pt}}m{#1}}
\newcolumntype{C}[1]{>{\centering\let\newline\\\arraybackslash\hspace{0pt}}m{#1}}

\usepackage[accsupp]{axessibility}  

\begin{document}
\pagestyle{headings}
\mainmatter
\def\ECCVSubNumber{7238}  

\title{Bi-directional Contrastive Learning for \\ Domain Adaptive Semantic Segmentation}


\titlerunning{Bi-directional Contrastive Learning for DASS}
%
\author{Geon Lee \and Chanho Eom \and Wonkyung Lee \and Hyekang Park \and Bumsub Ham\thanks{Corresponding author.} \\ \url{https://cvlab.yonsei.ac.kr/projects/DASS}}
\authorrunning{G. Lee et al.}
%
\institute{Yonsei University}
\maketitle

\begin{abstract}
  We present a novel unsupervised domain adaptation method for semantic segmentation that generalizes a model trained with source images and corresponding ground-truth labels to a target domain. A key to domain adaptive semantic segmentation is to learn domain-invariant and discriminative features without target ground-truth labels. To this end, we propose a bi-directional pixel-prototype contrastive learning framework that minimizes intra-class variations of features for the same object class, while maximizing inter-class variations for different ones, regardless of domains. Specifically, our framework aligns pixel-level features and a prototype of the same object class in target and source images (i.e., positive pairs), respectively, sets them apart for different classes (i.e., negative pairs), and performs the alignment and separation processes toward the other direction with pixel-level features in the source image and a prototype in the target image. The cross-domain matching encourages domain-invariant feature representations, while the bidirectional pixel-prototype correspondences aggregate features for the same object class, providing discriminative features. To establish training pairs for contrastive learning, we propose to generate dynamic pseudo labels of target images using a non-parametric label transfer, that is, pixel-prototype correspondences across different domains. We also present a calibration method compensating class-wise domain biases of prototypes gradually during training. Experimental results on standard benchmarks including GTA5 $\rightarrow$ Cityscapes and SYNTHIA $\rightarrow$ Cityscapes demonstrate the effectiveness of our framework.
  
\end{abstract}

\section{Introduction}
\label{sec:intro}
\begin{figure*}[!t]
  \begin{center}
  \includegraphics[width=0.9\linewidth]{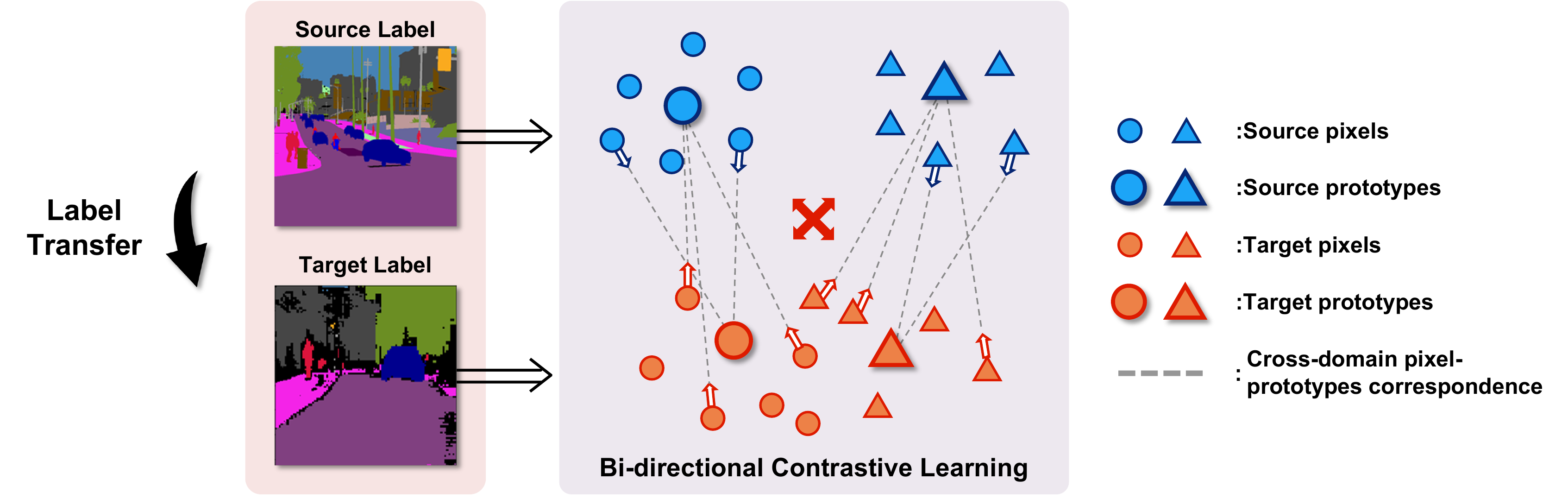}
     \caption{An illustration of our framework for UDASS. We generate pseudo labels of target images using a nonparametric label transfer. We then perform bi-directional pixel-prototype contrastive learning. This encourages pixel-level features in a target image and a prototype of the same object class in a source domain to pull each other, while setting them apart for different ones. We also perform the alignment and separation in a reverse direction, with pixel-level features of source images and prototypes of a target domain.}
  \label{fig:Teaser}
  \end{center}
  \vspace{-6mm}
\end{figure*}
\vspace{-3mm}

Semantic segmentation is to assign a semantic label to each pixel in an image. In the past decade, supervised methods based on convolutional neural networks (CNNs)~\cite{chen2014semantic,long2015fully,zheng2015conditional,ronneberger2015u,huang2019ccnet,song2019learnable} have achieved remarkable improvements in semantic segmentation. Training networks for the dense prediction task generally requires lots of pixel-level labels. Annotating pixel-level labels of high-resolution images is, however, significantly labor-intensive and time-consuming. For example, annotating the labels for an image of size $2048 \times 1024$ in Cityscapes~\cite{cordts2015cityscapes} takes about 90 minutes. One alternative is to leverage synthetic datasets,~\eg, GTA5~\cite{richter2016playing} and SYNTHIA~\cite{ros2016synthia}, that contain realistic images and corresponding pixel-level labels. The annotation cost is much cheaper than the manual labeling, but CNNs trained with synthetic datasets do not work well on real images, due to the domain discrepancy between synthetic and real images. 

To reduce the domain discrepancy, several methods~\cite{tsai2018learning,hoffman2018cycada,li2019bidirectional,chen2018road,hoffman2016fcns,tsai2019domain} have exploited an unsupervised domain adaptation approach. It transfers knowledge learned from a source domain (\eg,~a synthetic dataset) to a target one~(\eg,~a real dataset), with labels for the source domain alone. Many unsupervised domain adaptation methods leverage an adversarial training scheme~\cite{ganin2016domain} that aligns distributions of source and target domains by fooling a domain classifier~\cite{hoffman2018cycada,murez2018image,chen2019crdoco,li2019bidirectional,yang2020label,hoffman2016fcns,chen2017no,sankaranarayanan2017unsupervised,chen2018road,zhang2018fully,du2019ssf,luo2019significance,tsai2019domain,luo2019taking,sun2019not,wang2020differential}. However, they typically focus on reducing the domain discrepancy globally, and fail to keep pixel-level semantics~\cite{yang2020phase}. For example, regions corresponding to a car class in a source image might align with those for a bus class in a target image. Self-training methods~\cite{zou2018unsupervised,zou2019confidence,yang2020fda,yang2020phase,li2020content} enable a class-aware alignment. They generate pseudo labels for target images iteratively in a parametric approach, typically using CNNs trained with a source dataset, and then retrain a segmentation model on both source and target samples with the pseudo labels. This aligns cross-domain features in a class-level, improving performance of the model on target images progressively. The pseudo labels obtained using a parametric approach have the following drawbacks: First, they are very sparse, since low confident predictions are discarded to obtain reliable labels. Second, estimating pseudo labels is also computationally demanding, making them not to be updated frequently during training. These problems cause the segmentation model to overfit to pseudo labels, resulting in a large bias and a variance of predictions. In the following, we will call the labels estimated using a parametric approach as \emph{static pseudo labels}.

We present a novel contrastive learning framework using cross-domain pixel-prototype correspondences for unsupervised domain adaptive semantic segmentation~(UDASS). It aligns pixel-level features of each object class in target images, obtained by pseudo labels, with prototypes of corresponding class in a source domain, computed by ground truth, while setting them apart for different classes~(Fig.~\ref{fig:Teaser}). The alignment and separation process is also performed in a reverse direction, with pixel-level features of source images and prototypes of a target domain. The cross-domain matching encourages domain-invariant feature representations, and the bidirectional pixel-prototype correspondences provide compact and discriminative representations. We also present a nonparametric approach to generating \emph{dynamic pseudo labels} using pixel-prototype correspondences. Specifically, we calibrate prototypes of individual object classes in a source domain, while considering the domain discrepancy in target images, and establish correspondences for each prototype with individual pixel-level features in target images. We then transfer ground-truth labels of prototypes to corresponding pixels in target images. In contrast to the parametric approach in current self-training methods, our nonparametric approach provides denser pseudo labels, and generates the labels dynamically, whenever source images are changed during training. This helps to obtain more accurate pseudo labels, and prevents the overfitting problem. Experimental results on standard benchmarks including GTA5-to-Cityscapes~\cite{richter2016playing,cordts2015cityscapes} and SYNTHIA-to-Cityscapes~\cite{ros2016synthia,cordts2015cityscapes} demonstrate that our contrastive learning framework provides domain-invariant and discriminative features for UDASS. The main contributions can be summarized as follows:

\begin{itemize}
\item[$\bullet$] We introduce a novel contrastive learning framework using bi-directional pixel-prototype correspondences to learn domain-invariant and discriminative feature representations for UDASS.

\item[$\bullet$]  We propose a nonparametric approach to generating dynamic pseudo labels. We also present a calibration method to reduce domain biases for pixel-prototype correspondences between target and source domains.

\item[$\bullet$]  We set a new state of the art on standard benchmarks for UDASS, and demonstrate the effectiveness of our contrast learning framework.  
\end{itemize}

\vspace{-4mm}
\section{Related work}
\vspace{-2mm}
\subsubsection{UDASS.} 
UDASS leverages knowledge learned from a label-rich source domain to predict semantic labels of a scene in a target domain, where ground-truth annotations are not available. Synthetic images~(\eg, GTA5~\cite{richter2016playing} and SYNTHIA~\cite{ros2016synthia}) are widely used as source samples, as pixel-level labels can be generated automatically using computer graphics engines. The key factor for UDASS is hence to learn domain-invariant features to reduce the discrepancy between source and target domains. To this end, many UDASS methods adopt an adversarial learning framework~\cite{ganin2016domain} to fool a domain discriminator. They can generally be categorized into image-level and feature-level alignment methods. Motivated by image translation techniques~\cite{isola2017image,zhu2017cyclegan}, image-level alignment methods~\cite{hoffman2018cycada,chen2019crdoco,li2019bidirectional,murez2018image,yang2020label} transfer the styles~(\eg, texture and illumination) of target images to the source, so that segmentation models can accommodate both domains. Feature-level alignment methods~\cite{hoffman2016fcns,chen2017no,sankaranarayanan2017unsupervised,chen2018road,zhang2018fully,du2019ssf,luo2019significance,tsai2019domain,luo2019taking,sun2019not,wang2020differential} align the feature distributions of source and target images explicitly. These adversarial approaches, however, align source and target distributions globally. Namely, they perform a class-agnostic alignment, and ignore positional information of a scene. This suggests that the adversarial approaches fail to transfer pixel-level semantics, related to the structural information of a scene, from source to target domains.

UDASS methods based on self-training~\cite{zou2018unsupervised,zou2019confidence,li2020content} have recently been introduced. The self-training approach first segments target images using a model trained on a source dataset, and obtains pseudo labels if the confidence of semantic labels predicted by the model exceeds a pre-defined threshold. It then retrains the model iteratively with both ground-truth and pseudo labels of source and target datasets, respectively. The representative work of~\cite{zou2018unsupervised} proposes to use different thresholds for individual object categories to consider a class imbalance problem. In~\cite{zou2019confidence}, soft pseudo labels have been introduced, together with a confidence regularization technique that helps transfer discriminative feature representations from source to target domains. The self-training approaches~\cite{zou2018unsupervised,zou2019confidence,li2020content,zhang2021prototypical} are, however, likely to overfit to pseudo labels. The reasons are as follows:~(1) Pseudo labels are fixed for a few epochs during training, due to computational overheads, which accumulates error from incorrect pseudo labels;~(2) Pseudo labels are very sparse, as high confident predictions are chosen only as the labels. Our method alleviates these limitations by generating denser pseudo labels dynamically in a nonparametric way using pixel-prototype correspondences. Most similar to ours is PLCA~\cite{kang2020plca} using pixel-wise matches. It adopts a contrastive learning scheme to reduce the distances between source and target features directly at a pixel-level. The pixel-level domain alignment, however, does not consider contextual information, and fails to obtain compact representations between corresponding object categories in source and target domains. Our method instead uses bidirectional pixel-prototype correspondences for contrastive learning, which encourages intra-class compactness and inter-class separability across domains.

\vspace{-7mm}
\subsubsection{Prototypical learning.}
The seminal work of~\cite{snell2017prototypical} introduces prototypical networks that extract prototype representations for individual object categories. The prototypical features have proven useful in the limited-data regime for the task of,~\eg,~few-/zero-shot classification. PL~\cite{dong2018few} extends the idea of prototypical learning for few-shot semantic segmentation in such a way that class prototypes obtained from a support set are matched to pixel-level features in a query image. PANet~\cite{wang2019panet} presents a bidirectional framework exploiting correspondences between prototypical features for a support set and pixel-level ones for query images, and vice versa, for few-shot semantic segmentation. Similar to these methods, we exploit prototypical features for semantic segmentation. Differently, we leverage them within a framework of contrastive learning for UDASS. We use pixel-prototype correspondences to obtain domain-invariant and discriminative feature representations. We also leverage the correspondences to obtain dynamic pseudo labels, which alleviates the limitations of current self-training methods using static pseudo labels. 

\vspace{-3mm}
\subsubsection{Contrastive learning.} 
Contrastive learning~\cite{chen2020simple,he2020momentum} is a de facto approach to learning generic feature representations in a self-supervised way. The basic idea is to encourage positive pairs with the same label to be close, while negative ones with different labels to be distant. In order to set positive and negative pairs without ground-truth labels, contrastive learning augments a single input image,~\eg,~using random cropping and color jittering. It then considers the original image and the augmented one as a positive pair, while setting the pairs composed of the original and other images as negative ones. Similar to ours, CANet~\cite{kang2019contrastive} adopts contrastive learning for unsupervised domain adaptive classification. It computes the domain discrepancies using image-level features, and then performs a class-wise alignment using target labels obtained by a clustering method. Differently, our method leverages contrastive learning using correspondences between pixels and prototypes across domains. Optimizing bidirectional correspondences jointly in our method also enables aggregating features for the same object category, regardless of domains.

\vspace{-3mm}
\subsubsection{Nonparametric label transfer.} 
Label transfer has been widely used in object localization~\cite{malisiewicz2011ensemble}, scene segmentation~\cite{russell2009segmenting,liu2011nonparametric,najafi2016sample,singh2013nonparametric}, automatic image annotation~\cite{uricchio2017automatic}, and image translation~\cite{shrivastava2011data}. Label transfer methods first search visually similar images or patches in large datasets for given queries, and then transfer labels of retrieved samples to the queries. Similar to our approach, the work of~\cite{di2017cross} adopts a nonparametric label transfer method for scene parsing under different domains~(\eg, weather or illumination). Specifically, it extracts features from query images with pre-trained networks, finds the best matching images using SIFT flow~\cite{liu2010sift}, and transfers labels of the images to the queries via a probabilistic MRF model, suggesting that this approach requires source images and ground-truth labels at both training and test time. Our method, on the other hand, uses source images and corresponding ground-truth labels only at training time. Namely, we leverage non-parametric label transfer to train a parametric segmentation model. 

\begin{figure*}[t]
  \begin{center}
  \includegraphics[width=1\linewidth]{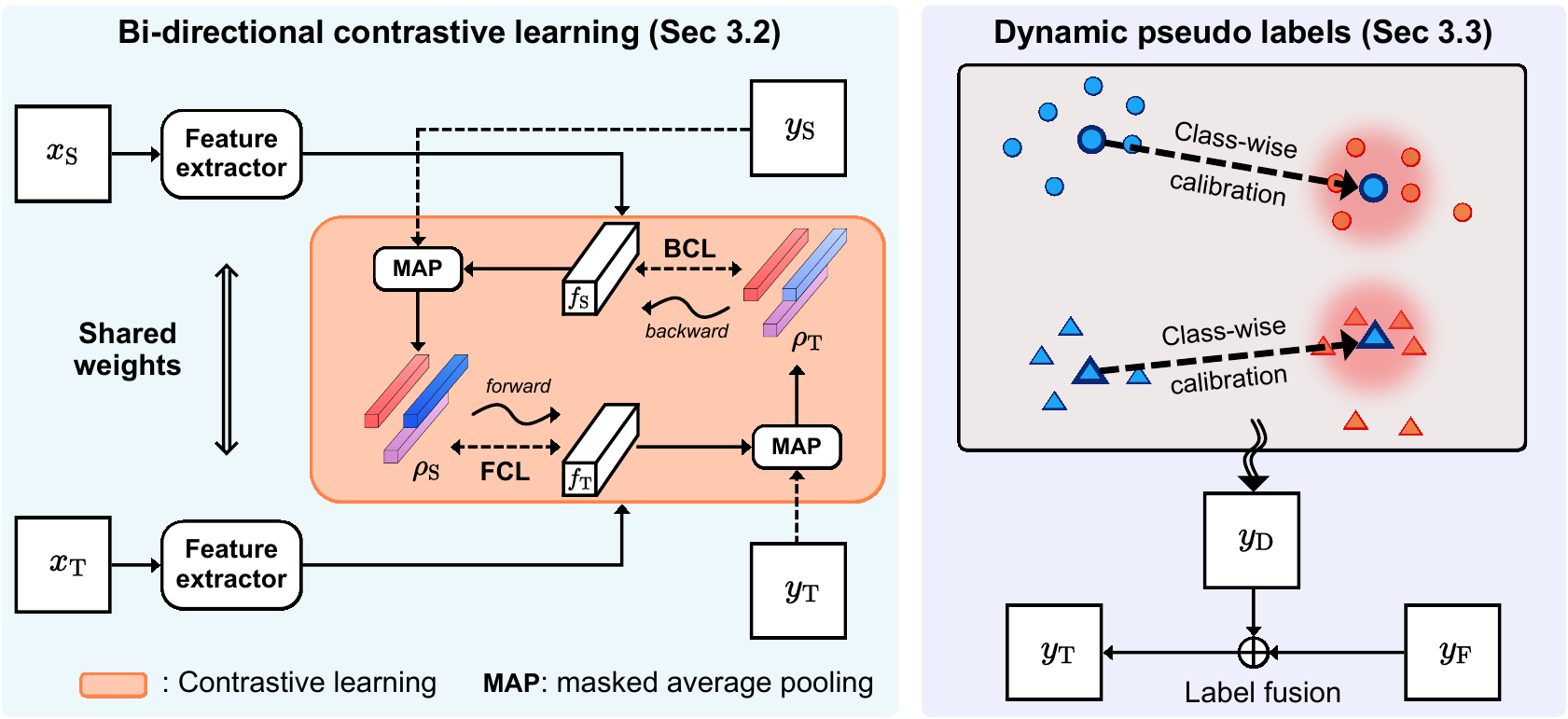}
     \caption{An overview of our framework. (Left) Bi-directional contrastive learning: We first extract feature maps, $f_\textrm{S}$ and $f_\textrm{T}$, from source and target images, $x_\textrm{S}$ and $x_\textrm{T}$, respectively. We then obtain prototypes in a source domain, $\rho_\textrm{S}$ using ground-truth labels of source images $y_{\textrm{S}}$. Prototypes in a target domain $\rho_\textrm{T}$ are similarly computed but with dynamic pseudo labels of target images $y_{\textrm{T}}$. Bidirectional contrastive terms, FCL and BCL, exploit pixel-prototype correspondences across domains to learn domain-invariant and discriminative features for UDASS. (Right) Hybrid pseudo labels: We generate dynamic pseudo labels~$y_{\textrm{D}}$ using pixel-prototype correspondences across domains, while calibrating the prototypes to reduce domain discrepancies. We then combine them with static ones~$y_{\textrm{F}}$ using a parametric approach to obtain hybrid pseudo labels~$y_{\textrm{T}}$.}
     \label{fig:Overview}
  \end{center}
  \vspace{-8mm}
\end{figure*}

\vspace{-3mm}
\section{Approach}
\vspace{-2mm}
\subsection{Overview}
\label{subsection:Overview}
We introduce a cross-domain contrastive learning framework for UDASS using pixel-prototype correspondences~(Fig.~\ref{fig:Overview}). It first extracts feature maps from source and target images, respectively, using a siamese network. We obtain prototypes of source and target domains using ground-truth labels of source images and pseudo labels of target ones, respectively. Our method then establishes correspondences between the prototypes and pixel-level features across domains, and leverages them to learn domain-invariant and discriminative representations via contrastive learning. To this end, we introduce a bi-directional contrastive loss that consists of a forward contrastive term~(FCL) and a backward contrastive term~(BCL). FCL matches individual pixel-level features of a target image with prototypes of a source domain, and enforces pixel-prototype pairs with the same class labels to be aligned closely than other ones. BCL performs the alignment process in a reverse direction, with pixel-level features of a source image and prototypes of a target domain, encouraging our model to provide discriminative and compact features. In order to establish training pairs for computing the bi-directional contrastive loss, we require pseudo labels of target images. To this end, we use dynamic pseudo labels obtained by a nonparametric label transfer, addressing the drawbacks of static pseudo labels. Specifically, given a pair of source-target images, we establish correspondences between prototypes of a source domain and pixel-level features of a target image, while calibrating the prototypes progressively during training to compensate domain discrepancies. We then set the pseudo labels of pixel-level features to the class labels of the corresponding prototypes in a source domain. Unlike static pseudo labels estimated by a parametric approach~\cite{zou2018unsupervised,li2020content}, our approach can generate novel pseudo labels of target images dynamically, whenever a pair of source-target images are changed, during training. We estimate hybrid pseudo labels by combining dynamic and static labels, and use them for the bi-directional contrastive learning.
\vspace{-3mm}
\subsection{Bi-directional contrastive learning}
\label{subsection:cross-domain}
Given a pair of source and target images, our goal is to aggregate pixel-level features for the same object class, regardless of domains, to learn domain-invariant and discriminative feature representations. To this end, we formulate UDASS as bi-directional pixel-prototype contrastive learning. Let us denote by $\mathcal{C}$ the set of object classes. We obtain prototypes of source and target domains for the class $c \in \mathcal{C}$, $\rho_\textrm{S}(c)$ and $\rho_\textrm{T}(c)$, using masked average pooling~(MAP) as follows: 
\begin{equation}
 \rho_\textrm{S}(c) =
         \frac{\sum_{p}f_\textrm{S}(p) y_\textrm{S}(p,c)} {\sum_{p} y_{\textrm{S}}(p,c)},
 \rho_\textrm{T}(c) =
         \frac{\sum_{p}f_\textrm{T}(p) y_\textrm{T}(p,c)} {\sum_{p} y_{\textrm{T}}(p,c)},	      
\label{eq:f_proto}
\end{equation}
where we denote by $f_\textrm{S}(p)$ and $f_\textrm{T}(p)$ pixel-level features of source and target images, respectively, at position $p$. $y_{\textrm{S}}(p,c)$ and $y_{\textrm{T}}(p,c)$ are one-hot labels,~\ie,~1 if the class label at position $p$ correspond to $c$ and 0 otherwise. Note that we use ground-truth labels of source images~$y_{\textrm{S}}$ and hybrid pseudo labels of target ones~$y_{\textrm{T}}$ to set the labels, $y_{\textrm{S}}$ and $y_{\textrm{T}}$, respectively. Using the prototypes of source and target domains, we perform cross-domain contrastive learning in a bi-directional way. We leverage a bi-directional constative loss that consists of FCL and BCL. FCL exploits prototypes of a source domain and pixel-level features of a target image. To be specific, given pixel-level features of a target image, we select the prototypes of a source domain having the same class labels as the features, and set them as positive pairs, while other prototypes are used to set negative ones. FCL maximizes the similarities between positive pairs as follows:
\begin{equation}
  \mathcal{L}_{FC} = - \sum_{c} \sum_{p} y_{\textrm{T}}(p,c) \log \frac{\exp \big({s(f_\textrm{T}(p)}, \rho_\textrm{S}(c)) / \tau \big)} {\sum_{c} \exp \big({s(f_\textrm{T}(p)}, \rho_\textrm{S}(c)) / \tau \big)},
\label{eq:L_fcl}
\end{equation}
where $\tau$ is a temperature parameter, and $s(\cdot, \cdot)$ computes cosine similarity. Similarly, BCL exploits prototypes of a target domain and pixel-level features of a source image. It encourages positive pairs sharing the same labels to pull each other, while making others set apart as follows: 
\begin{equation}
  \mathcal{L}_{BC} = - \sum_{c} \sum_{p} y_{\textrm{S}}(p,c) \log \frac{\exp \big({s(f_\textrm{S}(p)}, \rho_\textrm{T}(c)) / \tau \big)} {\sum_{c} \exp \big({s(f_\textrm{S}(p)}, \rho_\textrm{T}(c)) / \tau \big)}.
\label{eq:L_bcl}
\end{equation}
In summary, using the bidirectional contrastive loss, pixel-level features for the same object class are embedded closely, regardless of domains, while those for different classes are distinguished from each other. That is, by jointly optimizing FCL and BCL, we can minimize intra-class variations and maximize inter-class variations of pixel-level features progressively during training. In contrast to current UDASS methods~\cite{zhang2017curriculum,zou2018unsupervised,zou2019confidence,yang2020fda,yang2020phase,li2020content} that do not consider such variations for domain adaptation, our approach provides more discriminative and compact features. This in turn allows to perform more accurate class-wise alignments across domains, and enables our model to generalize better on a target domain.
\vspace{-3mm}
\subsection{Dynamic pseudo labels}
\label{subsection:Dynamic pseudo labels}
\begin{figure}[t]
  \begin{center}
  \includegraphics[width=0.7\linewidth]{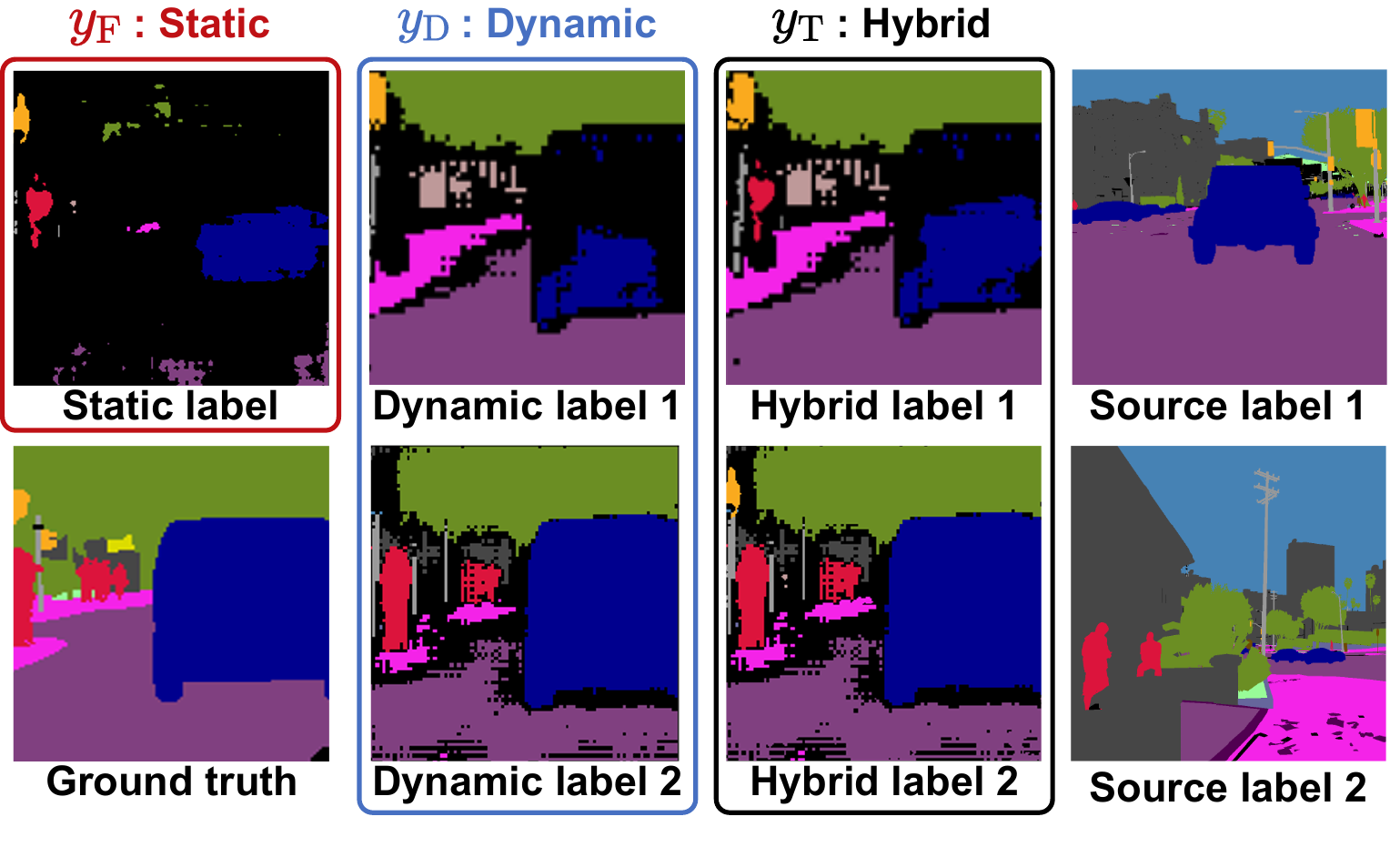}
     \vspace{-2mm}
     \caption{Visual comparison of static, dynamic, and hybrid pseudo labels for a target image. In contrast to the static label~(the first column), estimated using a parametric segmentation model, dynamic labels are obtained by a nonparametric label transfer between source and target images. This suggests that they are denser and cheap to update. We can also obtain different dynamic labels~(the second column), according to source images~(the fourth column). We combine both labels to get hybrid pseudo labels~(the third column), and use them to augment the number of positive and negative pairs for contrastive learning.}
     \label{fig:label fusion dynamic}
  \end{center}
  \vspace{-6mm}
\end{figure}

Current self-training methods~\cite{zhang2017curriculum,zou2018unsupervised,zou2019confidence,yang2020fda,yang2020phase,li2020content} employ a parametric model trained with ground-truth labels of source images to obtain static pseudo labels of target images. Specifically, using the parametric segmentation model, confidence scores for individual object classes are computed for each pixel-level feature from entire target images. The pixel-level features with high confidence scores are chosen, and corresponding object classes are used as pseudo labels. Although exploiting static pseudo labels of target images enables performing class-aware UDASS, they have the following drawbacks: First, computing the pixel-level confidence scores for all target images to obtain the pseudo labels is computationally demanding. Current self-training methods perform this process for a few iterations~(\eg,~10000) during training, and update the pseudo labels of target images very occasionally. The error from incorrect pseudo labels might hence be accumulated. Second, current self-training methods choose highly confident pixel-level features only for static pseudo labels, and thus they are very sparse. These problems cause a model to overfit to the static pseudo labels, and induce suboptimal class-wise alignments between domains. To overcome the limitations, we introduce a novel approach to generating dynamic pseudo labels. It leverages a nonparametric label transfer technique using pixel-prototype correspondences between source and target images. That is, we estimate pseudo labels using pairs of source and target images. This suggests that our approach generates pseudo labels of target images dynamically, whenever source images are changed during training. In other words, the pseudo labels for the same target image could be different, depending on which source images are used to establish pixel-prototype correspondences w.r.t the target one~(Fig.~\ref{fig:label fusion dynamic}).

Concretely, given a pair of source and target images, we establish correspondences between prototypes of a source image and pixel-level features of a target one. To obtain reliable correspondences, we alleviate domain biases between source and target domains. We could estimate the degree of domain biases by calculating average class-wise features for each domain using all source and target images, followed by computing differences between the average features, which however requires lots of computational overheads. We instead leverage prototypes of source and target images. We first update prototypes of source and target domains progressively during training using an exponential moving average with a momentum parameter of~$\lambda$ as follows: 
\begin{align}
 \mu_\textrm{S}(c) \leftarrow \lambda \mu_\textrm{S}(c) + (1-\lambda)\rho_\textrm{S}(c), \\
  \mu_\textrm{T}(c) \leftarrow \lambda \mu_\textrm{T}(c) + (1-\lambda)\rho_\textrm{T}(c),
\end{align} 
where we denote by $\mu_\textrm{S}(c)$ and $\mu_\textrm{T}(c)$ updated prototypes of source and target domains, respectively, for the class~$c$.
We then estimate class-wise domain biases:
\begin{equation}
  \xi(c) = \mu_\textrm{T}(c) - \mu_\textrm{S}(c), \\
\end{equation}
and obtain calibrated prototypes for each object class in a source domain as follows:
\begin{equation}
\rho_{\textrm{S} \to \textrm{T}}(c) = \rho_\textrm{S}(c) + \xi(c).	
\end{equation}
Using the calibrated prototypes, we can establish more correct correspondences across domains. We consider the correspondences are correct, if similarity scores between the pixel-prototype matches are larger than a pre-defined threshold, and set dynamic pseudo labels of target images to corresponding object categories of the prototypes, as follows:  
\begin{equation}
  y_{\textrm{D}}(p,c) = 
\begin{cases}
 1,& \text{if  } s(f_\textrm{T}(p)),  \rho_{\textrm{S} \to \textrm{T}}(c)) > \mathcal{T} \text{and } c=c^\prime\\
 0,& \text{otherwise} \\ 
\end{cases},
\label{eq:dynamic pseudo label}
\end{equation}
where $y_{\textrm{D}}(p,c)$ is a dynamic pseudo label for the class $c$ at position $p$, $\mathcal{T}$ is a pre-defined threshold, and 
\begin{equation}
 c^\prime = \argmax_c(s(f_\textrm{T}(p)),  \rho_{\textrm{S} \to \textrm{T}}(c))).
\label{eq:argmax}
\end{equation}

\vspace{-7mm}
\subsubsection{Hybrid pseudo labels.}
We can obtain diverse pseudo labels even for the same target image every iteration, and the dynamic labels are much denser than static ones. Static pseudo labels, on the other hand, are sparse but reliable. In order to take advantage of both, we combine them and obtain hybrid pseudo labels~$y_{\textrm{T}}$ as follows:  
\begin{equation}
  y_{\textrm{T}}(p,c) = 
\begin{cases}
  y_{\textrm{D}}(p,c),& \text{if  } y_{\textrm{D}}(p,c) = 1 \\
  y_{\textrm{F}}(p,c),& \text{if  } y_{\textrm{D}}(p,c^\prime) = 0 \text{   for  } c^\prime \in \mathcal{C} \text{, and     } y_{\textrm{F}}(p,c) = 1 \\
  0,& \text{otherwise} \\ 
\end{cases},
\label{eq:label fusion}
\end{equation}
where $y_{\textrm{F}}(p,c)$ is a static pseudo label for the class $c$ at position $p$.

\vspace{-3mm}
\subsection{Training loss}
Following the previous works~\cite{zou2018unsupervised,zou2019confidence,yang2020fda,yang2020phase,li2020content}, we exploit segmentation and entropy terms using ground-truth and pseudo labels of source and target images, respectively. The former encourages our model to provide accurate pixel-wise predictions, and the latter minimizes the entropy of the predictions. We define a loss for training a baseline model as follows:
\begin{equation}
  \mathcal{L}_\textrm{base} = \lambda_{seg}^{\textrm{S}} \mathcal{L}_{seg}^{\textrm{S}} + \lambda_{seg}^{\textrm{T}}\mathcal{L}_{seg}^{\textrm{T}} +  \lambda_{ent}^{\textrm{S}}\mathcal{L}_{ent}^{\textrm{S}} + \lambda_{ent}^{\textrm{T}}\mathcal{L}_{ent}^{\textrm{T}}, 
  \label{eq:L_baseline}
\end{equation}
where $\mathcal{L}_{seg}^{\textrm{S}}$ and $\mathcal{L}_{seg}^{\textrm{T}}$ are segmentation losses for source and target domains, respectively. $\mathcal{L}_{ent}^{\textrm{S}}$ and $\mathcal{L}_{ent}^{\textrm{T}}$ are entropy terms for source and target domains, respectively. $\lambda_{seg}^{\textrm{S}}$, $\lambda_{seg}^{\textrm{T}}$, $\lambda_{ent}^{\textrm{S}}$, and $\lambda_{ent}^{\textrm{T}}$ are balance parameters for each term. For the baseline, we obtain static pseudo labels using the method of~\cite{zou2018unsupervised}. As our final model, we additionally use a bi-directional contrastive loss to learn domain-invariant and discriminative representations as follows: 
\begin{equation}
  \mathcal{L} = \mathcal{L}_\textrm{base} + \lambda_{FC}\mathcal{L}_{FC} + \lambda_{BC}\mathcal{L}_{BC},
\label{eq:Total loss}
\end{equation}
where $\lambda_{FC}$ and $\lambda_{BC}$ are weighting factors for forward and backward contrastive terms, respectively.

\vspace{-4mm}
\section{Experiments}
\vspace{-2mm}
\subsection{Implementation details}
\subsubsection{Dataset and evaluation metric.} We evaluate our framework on two standard benchmarks (GTA5~\cite{richter2016playing} $\rightarrow$ Cityscapes~\cite{cordts2015cityscapes}, and SYNTHIA~\cite{ros2016synthia} $\rightarrow$ Cityscapes~\cite{cordts2015cityscapes}). GTA5 and SYNTHIA provide 24,996 and 9400 images, respectively. Cityscapes consists of 2975, 500, and 1525 images for training, validation, and testing, respectively. Following the standard protocol in \cite{hoffman2018cycada,tsai2018learning,zou2018unsupervised,zou2019confidence}, we report the mean intersection over Union (mIoU) on 19 classes for GTA5 $\rightarrow$ Cityscapes and 13 (or 16) classes for Synthia $\rightarrow$ Cityscapes.
  
\vspace{-3mm}
\subsubsection{Training.} We adopt the DeepLab-V2~\cite{chen2014semantic} architecture with ResNet-101~\cite{he2016deep} as a backbone network pretrained for ImageNet classification~\cite{deng2009imagenet}. We first train DeepLab-V2 with a source dataset, and use it as an initial segmentation model for UDASS. 
We train the model for 100$k$ iterations with a batch size of 4, using stochastic gradient descent~(SGD)~\cite{kiefer1952stochastic} of a momentum of 0.9 and weight decay of $5 \times 10^{-4}$. We use a poly learning rate scheduling with an initial learning rate of $7.5 \times 10^{-5}$. We update static pseudo labels $y_{\textrm{F}}$ every 10$k$ iterations. We resize a shorter side of images to 850, and crop them into a patch of size~730 $\times$ 730. For data augmentation, we use horizontal flipping and random scaling with the factor of [0.8, 1.2]. We use a weighted sampling strategy to select source images containing objects that rarely appear in a source domain, mitigating low co-occurrence rates for the rare object categories. Following \cite{zhang2021prototypical,fang2021seed}, we additionally apply a self-distillation technique to our final model.
 Detailed descriptions for the weighted sampling and hyperparameter settings are available in the supplement.

\begin{table*}[t]
   \centering
   \resizebox{0.99\linewidth}{!}{
    \large
   \begin{tabular}{clcc@{\hspace{0.2cm}}c@{\hspace{0.2cm}}c@{\hspace{0.2cm}}c@{\hspace{0.2cm}}c@{\hspace{0.2cm}}c@{\hspace{0.2cm}}c@{\hspace{0.2cm}}c@{\hspace{0.2cm}}c@{\hspace{0.2cm}}c@{\hspace{0.2cm}}c@{\hspace{0.2cm}}c@{\hspace{0.2cm}}c@{\hspace{0.2cm}}c@{\hspace{0.2cm}}c@{\hspace{0.2cm}}c@{\hspace{0.2cm}}c@{\hspace{0.2cm}}c@{\hspace{0.2cm}}c@{\hspace{0.2cm}}c@{\hspace{0.2cm}}}
   \hline
   \multicolumn{23}{c}{GTA5 $\rightarrow$ Cityscapes}                                                                    \\ \hline
   \multicolumn{1}{l@{}}{\rotatebox{90}{Split}}  & \multicolumn{1}{c@{}}{Methods}                & \multicolumn{1}{@{\hspace{0.2cm}}c@{\hspace{0.2cm}}}{\rotatebox{90}{Type   }} & \rotatebox{90}{road} & \rotatebox{90}{side.} & \rotatebox{90}{build.} & \rotatebox{90}{wall} & \rotatebox{90}{fence} & \rotatebox{90}{pole} & \rotatebox{90}{light} & \rotatebox{90}{sign} & \rotatebox{90}{veg.} & \rotatebox{90}{terrian} & \rotatebox{90}{sky}  & \rotatebox{90}{person} & \rotatebox{90}{rider} & \rotatebox{90}{car}  & \rotatebox{90}{truck} & \rotatebox{90}{bus}  & \rotatebox{90}{train} & \rotatebox{90}{motor} & \rotatebox{90}{bike} & \multicolumn{1}{c}{\rotatebox{90}{mIoU   }} \\ 
   \hline
   \multicolumn{1}{l}{\multirow{12}{*}{\rotatebox{90}{Validation}}} &\multicolumn{1}{l@{}}{Source-only}        & \multicolumn{1}{@{}c@{}}{-}     & 45.4 & 16.5  & 66.4   & 14.4 & 21.6  & 25.1 & 36.3  & 17.2 & 80.1 & 16.3    & 69.1 & 61.4   & 24.9  & 68.6 & 28.4  & 4.7 & 4.4   & 40.8 & 27.5 & \multicolumn{1}{@{}c}{35.2} \\ 
   \multicolumn{1}{l}{} & \multicolumn{1}{l@{}}{AdaptSeg~\cite{tsai2018learning}}        & \multicolumn{1}{@{}c@{}}{AT}     & 86.5 & 36.0  & 79.9   & 23.4 & 23.3  & 23.9 & 35.2  & 14.8 & 83.4 & 33.3    & 75.6 & 58.5   & 27.6  & 73.7 & 32.5  & 35.4 & 3.9   & 30.1  & 28.1 & \multicolumn{1}{@{}c}{42.4} \\
   \multicolumn{1}{l}{} & \multicolumn{1}{l@{}}{CBST~\cite{zou2018unsupervised}}            & \multicolumn{1}{@{}c@{}}{ST}     & 91.8 & 53.5  & 80.5   & 32.7 & 21.0  & 34.0 & 28.9  & 20.4 & 83.9 & 34.2    & 80.9 & 53.1   & 24.0  & 82.7 & 30.3  & 35.9 & 16.0  & 25.9  & 42.8 & \multicolumn{1}{@{}c}{45.9}\\
   \multicolumn{1}{l}{} & \multicolumn{1}{l@{}}{CRST~\cite{zou2019confidence}}            & \multicolumn{1}{@{}c@{}}{ST}     & 91.0 & 55.4  & 80.0   & 33.7 & 21.4  & 37.3 & 32.9  & 24.5 & 85.0 & 34.1    & 80.8 & 57.7   & 24.6  & 84.1 & 27.8  & 30.1 & 26.9  & 26.0  & 42.3 & \multicolumn{1}{@{}c}{47.1}\\
   \multicolumn{1}{l}{} & \multicolumn{1}{l@{}}{PLCA~\cite{kang2020plca}}    & \multicolumn{1}{@{}c@{}}{-}      & 84.0 & 30.4  & 82.4   & 35.3 & 24.8  & 32.2 & 36.8  & 24.5 & 85.5 & 37.2    & 78.6 & 66.9   & 32.8  & 85.5 & 40.4  & 48.0 & 8.8   & 29.8  & 41.8 & \multicolumn{1}{@{}c}{47.7}\\
   \multicolumn{1}{l}{} & \multicolumn{1}{l@{}}{CAG\_UDA~\cite{zhang2019category}}        & \multicolumn{1}{@{}c@{}}{ST}     & 90.4 & 51.6  & 83.8   & 34.2 & 27.8  & 38.4 & 25.3  & 48.4 & 85.4 & 38.2    & 78.1 & 58.6   & 34.6  & 84.7 & 21.9  & 42.7 & \textbf{41.1}  & 29.3  & 37.2 & \multicolumn{1}{@{}c}{50.2} \\
   \multicolumn{1}{l}{} & \multicolumn{1}{l@{}}{FDA~\cite{yang2020fda}}        & \multicolumn{1}{@{}c@{}}{ST}     & 92.5 & 53.5  & 82.4   & 26.5 & 27.6  & 36.4 & 40.6  & 38.9 & 82.3 & 39.8    & 78.0 & 62.6   & 34.4  & 84.9 & 34.1  & 53.1 & 16.9  & 27.7  & 46.4 & \multicolumn{1}{@{}c}{50.5} \\
   \multicolumn{1}{l}{} & \multicolumn{1}{l@{}}{TPLD~\cite{shin2020two}}        & \multicolumn{1}{@{}c@{}}{ST}     & 94.2 & 60.5  & 82.8   & 36.6 & 16.6  & 39.3 & 29.0  & 25.5& 85.6 & 44.9    & 84.4 & 60.6   & 27.4  & 84.1 & 37.0  & 47.0 & 31.2  & 36.1  & 46.4 & \multicolumn{1}{@{}c}{51.2} \\
   \multicolumn{1}{l}{} & \multicolumn{1}{l@{}}{CorDA~\cite{wang2021domain}}        & \multicolumn{1}{@{}c@{}}{ST}     & \textbf{94.7} & \textbf{63.1}  & 87.6   & 30.7 & 40.6  & 40.2 & 47.8  & 51.6 & 87.6 & \textbf{47.0}   & 89.7 & 66.7   & 35.9  & 90.2 & \textbf{48.9}  & 57.5 & 0.0  & 39.8  & 56.0 & \multicolumn{1}{@{}c}{56.6} \\
   \multicolumn{1}{l}{} & \multicolumn{1}{l@{}}{ProDA~\cite{zhang2021prototypical}}       & \multicolumn{1}{@{}c@{}}{ST}     & 87.1 &55.1 &78.1 &\textbf{45.6} &\textbf{43.8} &\textbf{44.6} &52.5 &\textbf{53.4} & 89.1 &44.7 &82.1 &70.1 &\textbf{39.1} &88.4 &43.8 &\textbf{59.1} &1.0 &\textbf{48.7} &  54.4 & \multicolumn{1}{@{}c}{56.5} \\ 
    \multicolumn{1}{l}{} & \multicolumn{1}{l@{}}{Ours}            & \multicolumn{1}{@{}c@{}}{ST}  &  93.5   &  60.2    &   \textbf{88.1}    &    31.1    &   37.0   &  41.9     &  \textbf{54.7}    &   37.8    &  \textbf{89.9}    &  45.5    &    \textbf{89.9}     &  \textbf{72.7}    &    38.2    &   \textbf{90.7}    &   34.3   &   53.2    &   4.4   &   47.2        & \textbf{58.5}     &   \multicolumn{1}{@{}c}{\textbf{57.1}  }  \\ \hline 

   \multicolumn{1}{l}{\multirow{7}{*}{\rotatebox{90}{Test}}} & \multicolumn{1}{l@{\hspace{0.2cm}}}{AdaptSeg~\cite{tsai2018learning}} & \multicolumn{1}{@{}c@{}}{AT}     & 88.5 & 40.4  & 81.0   & 26.3 & 20.6  & 25.6 & 36.0  & 12.9 & 84.8 & 45.5    & 87.2 & 63.7   & 35.8  & 76.4 & 27.7  & 28.0 & 2.9  & 33.0  & 26.1 & \multicolumn{1}{@{}c}{44.3}  \\ 
   \multicolumn{1}{l}{} & \multicolumn{1}{l@{}}{CBST~\cite{zou2018unsupervised}}            & \multicolumn{1}{@{}c@{}}{ST}     & 91.0 & 55.4  & 80.0   & 33.7 & 21.4  & 37.3 & 32.9  & 24.5 & 85.0 & 34.1    & 80.8 & 57.7   & 24.6  & 84.1 & 27.8  & 30.1 & \textbf{26.9}  & 26.0  & 42.3 &  \multicolumn{1}{@{}c}{47.1}  \\
   \multicolumn{1}{l}{} & \multicolumn{1}{l@{}}{CRST~\cite{zou2019confidence}}             & \multicolumn{1}{@{}c@{}}{ST}     & 93.5 & 57.6  & 84.6   & \textbf{39.3} & 24.1  & 25.2 & 35.0  & 17.3 & 85.0 & 40.6    & 86.5 & 58.7   & 28.7  & 85.8 & \textbf{49.0}  & 56.4 & 5.4   & 31.9  & 43.2 & \multicolumn{1}{@{}c}{49.9} \\
   \multicolumn{1}{l}{} & \multicolumn{1}{l@{}}{FDA-MBT~\cite{yang2020fda}}             & \multicolumn{1}{@{}c@{}}{ST}  & 93.4 & 55.8 & 83.6 & 25.4 & 23.1 & 33.2 & 39.0 & 36.9 & 84.0 & 47.2 & 88.8 & 66.3 & 40.6 & 87.4 & 26.9 & 49.6 & 12.8 & 35.2 & 42.8 & \multicolumn{1}{@{}c}{51.2} \\
   \multicolumn{1}{l}{} & \multicolumn{1}{l@{}}{CorDA~\cite{wang2021domain}}        & \multicolumn{1}{@{}c@{}}{ST}     & \textbf{94.2} & \textbf{62.9}  & 88.1   & 30.2 & 41.2  & 40.1 & 49.1  & 49.9 & 89.1 & 49.1  & 90.1 & 69.1   & 28.9  & 86.2 & 46.2  & 59.5 & 1.2  & 35.2  & 52.3 & \multicolumn{1}{@{}c}{57.5} \\
   \multicolumn{1}{l}{} & \multicolumn{1}{l@{}}{ProDA~\cite{zhang2021prototypical}}        & \multicolumn{1}{@{}c@{}}{ST}     & 88.1 &57.1 &81.2 & 46.1 & \textbf{45.2} & \textbf{41.5} &\textbf{55.1} & \textbf{56.2} &86.1 &45.1 & 78.1 &73.2 &40.1 &88.8 &48.7 &\textbf{60.1} &1.1 &50.3 & 53.1 & \multicolumn{1}{@{}c}{57.6}  \\
   \multicolumn{1}{l}{} & \multicolumn{1}{l@{}}{Ours}            & \multicolumn{1}{@{}c@{}}{ST} & 93.8 & 59.7 & \textbf{90.1} & 38.0 & 33.4 & 39.9 & 45.3 & 30.5 & \textbf{92.2} & \textbf{58.2} & \textbf{94.8} & \textbf{81.9} & \textbf{47.9} & \textbf{93.2} & 40.1 & 53.1 & 13.1 & \textbf{51.2} & \textbf{58.2}  & \multicolumn{1}{@{}c}{ \textbf{58.5}  } \\ \hline
   \end{tabular}
   }
   \vspace{2mm}
   \captionsetup{font={small}}
   \caption{Quantitative comparison with state-of-the-art methods on GTA5 $\rightarrow$ Cityscapes in terms of mIoU. AT: methods based on adversarial training; ST: methods based on self-training. $\dagger$: a method using a different network architecture.}
   \vspace{-2mm}
   \label{tab:gta}
   \end{table*}

   \begin{table*}[t]
      \centering
      \resizebox{0.99\linewidth}{!}{
        \large
      \begin{tabular}{lcc@{\hspace{0.2cm}}c@{\hspace{0.2cm}}c@{\hspace{0.2cm}}c@{\hspace{0.2cm}}c@{\hspace{0.2cm}}c@{\hspace{0.2cm}}c@{\hspace{0.2cm}}c@{\hspace{0.2cm}}c@{\hspace{0.2cm}}c@{\hspace{0.2cm}}c@{\hspace{0.2cm}}c@{\hspace{0.2cm}}c@{\hspace{0.2cm}}c@{\hspace{0.2cm}}c@{\hspace{0.2cm}}c@{\hspace{0.2cm}}@{\hspace{0.2cm}}c@{\hspace{0.2cm}}c}
      \hline
      \multicolumn{20}{c}{SYNTHIA $\rightarrow$ Cityscapes}                                                                                                                                                                             \\ \hline
      \multicolumn{1}{c@{}}{Methods}                & \multicolumn{1}{@{\hspace{0.2cm}}c}{\rotatebox{90}{Type   }} & \rotatebox{90}{road} & \rotatebox{90}{side.} & \rotatebox{90}{build.} & \rotatebox{90}{wall*} & \rotatebox{90}{fence*} & \rotatebox{90}{pole*} & \rotatebox{90}{light} & \rotatebox{90}{sign} & \rotatebox{90}{veg.} & \rotatebox{90}{sky}  & \rotatebox{90}{person} & \rotatebox{90}{rider} & \rotatebox{90}{car}  & \rotatebox{90}{bus}  & \rotatebox{90}{motor} & \rotatebox{90}{bike} & \rotatebox{90}{mIoU} & \rotatebox{90}{mIoU*} \\ \hline
      \multicolumn{1}{l@{}}{Source-only}        & \multicolumn{1}{@{}c@{}}{AT}     & 53.4 & 23.4  & 73.0   & 5.5    & 0.0      & 25.7     & 6.6   & 7.0  & 77.9 & 55.3 & 52.9   & 21.0  & 60.9 & 6.6 & 21.8 & 33.7 & 32.5    &\multicolumn{1}{@{}c}{37.6}  \\ 
      \multicolumn{1}{l@{}}{AdaptSeg~\cite{tsai2018learning}}        & \multicolumn{1}{@{}c@{}}{AT}     & 84.3 & 42.7  & 77.5   & -     & -      & -     & 4.7   & 7.0  & 77.9 & 82.5 & 54.3   & 21.0  & 72.3 & 32.2 & 18.9  & 32.3 & -    & \multicolumn{1}{@{}c}{46.7} \\
      \multicolumn{1}{l@{}}{CBST~\cite{zou2018unsupervised}}            & \multicolumn{1}{@{}c@{}}{ST}     & 68.0 & 29.9  & 76.3   & 10.8  & 1.4    & 33.9  & 22.8  & 29.5 & 77.6 & 78.3 & 60.6   & 28.3  & 81.6 & 23.5 & 18.8  & 39.8 & 38.9 & \multicolumn{1}{@{}c}{42.6 } \\
      \multicolumn{1}{l@{}}{CRST~\cite{zou2019confidence}}            & \multicolumn{1}{@{}c@{}}{ST}     & 67.7 & 32.2  & 73.9   & 10.7  & 1.6    & 37.4  & 22.2  & 31.2 & 80.8 & 80.5 & 60.8   & 29.1  & 82.8 & 25.0 & 19.4  & 45.3 & 43.8 & \multicolumn{1}{@{}c}{50.1 } \\
      \multicolumn{1}{l@{}}{CAG\_UDA~\cite{zhang2019category}}        & \multicolumn{1}{@{}c@{}}{ST}     & 84.7 & 40.8  & 81.7   & 7.8   & 0.0    & 35.1  & 13.3  & 22.7 & 84.5 & 77.6 & 64.2   & 27.8  & 80.9 & 19.7 & 22.7  & 48.3 & 44.5 & \multicolumn{1}{@{}c}{51.5 } \\
      \multicolumn{1}{l@{}}{FDA~\cite{yang2020fda}}        & \multicolumn{1}{@{}c@{}}{ST}    & 79.3 & 35.0 & 73.2 & - & - & - & 19.9 & 24.0 & 61.7 & 82.6 & 61.4 & 31.1 & 83.9 & 40.8 & 38.4  & 51.1 & - & \multicolumn{1}{@{}c}{52.5} \\
      \multicolumn{1}{l@{}}{PLCA~\cite{kang2020plca}}    & \multicolumn{1}{@{}c@{}}{-}      & 82.6 & 29.0  & 81.0   & 11.2  & 0.2    & 33.6  & 24.9  & 18.3 & 82.8 & 82.3 & 62.1   & 26.5  & 85.6 & 48.9 & 26.8  & 52.2 & 46.8 & \multicolumn{1}{@{}c}{54.0}  \\ 
      \multicolumn{1}{l@{}}{TPLD~\cite{shin2020two}}        & \multicolumn{1}{@{}c@{}}{ST}    & 80.9 & 44.3 & 82.2 & 19.9 & 0.3 & 40.6 & 20.5 & 30.1 & 77.2 & 80.9 & 60.6 & 25.5 & 84.8 & 41.1& 24.7  & 43.7 & 47.3 & \multicolumn{1}{@{}c}{53.5}  \\
      \multicolumn{1}{l@{}}{CorDA~\cite{wang2021domain}}        & \multicolumn{1}{@{}c@{}}{ST}  & \textbf{93.3} &  \textbf{61.6} &  85.3 &  19.6 &  5.1 &  37.8 &  36.6 &  \textbf{42.8} &  84.9 &  90.4 &  69.7 &  41.8 &  85.6 & 38.4 &  32.6 & 53.9 &  55.0 &  \multicolumn{1}{@{}c}{62.8} \\
      \multicolumn{1}{l@{}}{ProDA~\cite{zhang2021prototypical}}        & \multicolumn{1}{@{}c@{}}{ST} & 87.3 & 45.1 & 84.2 & \textbf{36.5} & 0.0 & \textbf{43.3} & \textbf{54.7} & 36.0 & 88.3 & 83.1 & 71.5 & 24.4 & 88.4 & \textbf{50.1} & 40.1 & 45.6 &  55.1 & \multicolumn{1}{@{}c}{61.3} \\
      \multicolumn{1}{l@{}}{Ours}            & \multicolumn{1}{@{}c@{}}{ST}      & 83.8 & 42.2  &  \textbf{85.3}  & 16.4  &  \textbf{5.7}   &  43.1 &  48.3 & 30.2 & \textbf{89.3} & \textbf{92.1} & \textbf{68.2}  & \textbf{43.1} & \textbf{89.7} & 47.2 &  \textbf{42.2} & \textbf{54.2} & \textbf{55.6} & \multicolumn{1}{@{}c}{\textbf{62.9}} \\ \hline
      \end{tabular}
      }
      \vspace{2mm}
      \captionsetup{font={small}}
      \caption{Quantitative comparison with state-of-the-art methods on SYNTHIA $\rightarrow$ Cityscapes results in terms of mIoU. We report the results for 13 classes~(mIoU$^*$) and 16 classes~(mIoU). AT: methods based on adversarial training; ST: methods based on self-training.}
      \vspace{-6mm}
      \label{tab:synthia}
   \end{table*}

\begin{figure*}[t]
   \begin{center}    
   \begin{subfigure}{0.24\textwidth}
      \centering
      \includegraphics[width=1\linewidth]{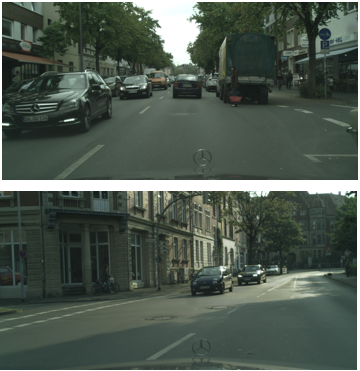}
      \captionsetup{font={small}}   
      \caption{Target images.}
   \end{subfigure}
   \begin{subfigure}{0.24\textwidth}
      \centering
      \includegraphics[width=1\linewidth]{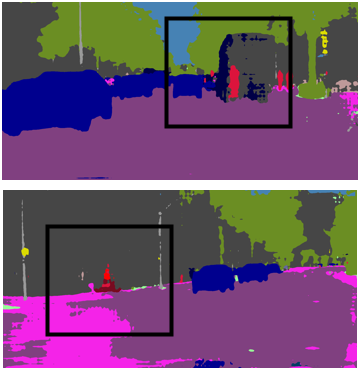}
      \captionsetup{font={small}}
      \caption{Baseline.}
   \end{subfigure}      
   \begin{subfigure}{0.24\textwidth}
      \centering
      \includegraphics[width=1\linewidth]{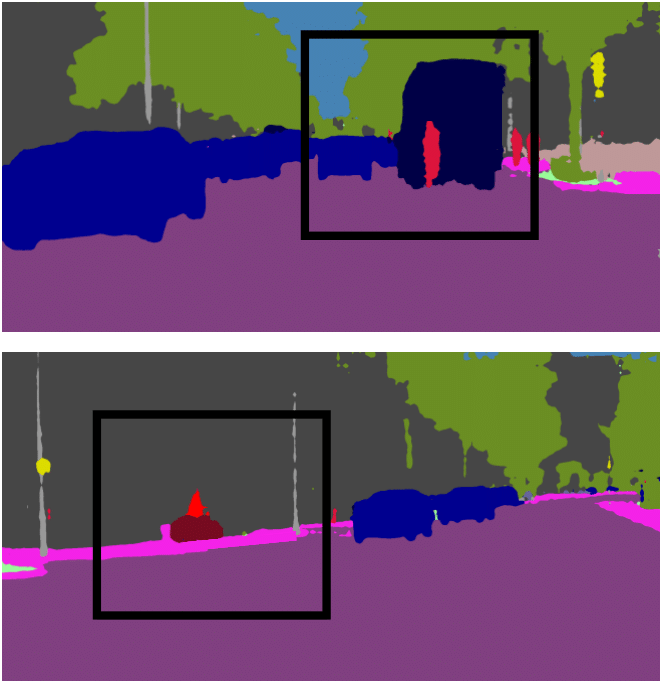}
      \captionsetup{font={small}}
      \caption{Ours.}
   \end{subfigure}
   \begin{subfigure}{0.24\textwidth}
      \centering
      \includegraphics[width=1\linewidth]{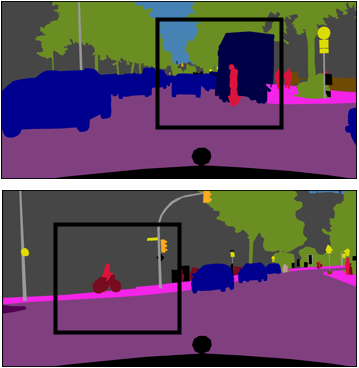}
      \captionsetup{font={small}}
      \caption{GT labels.}
   \end{subfigure}    
    \captionsetup{font={small}}
   \caption{Qualitative comparisons on GTA5 $\rightarrow$ Cityscapes. Our model gives better results than the baseline. (Best viewed in color)}
   \label{fig:segmentation}
\end{center}
\vspace{-2mm}
\end{figure*}


\renewcommand{\arraystretch}{1}

\begin{figure}[t]
  \begin{minipage}{0.5\linewidth}
  	  \centering
      \begin{subfigure}{0.45\textwidth}
        \centering
        \includegraphics[width=1\linewidth]{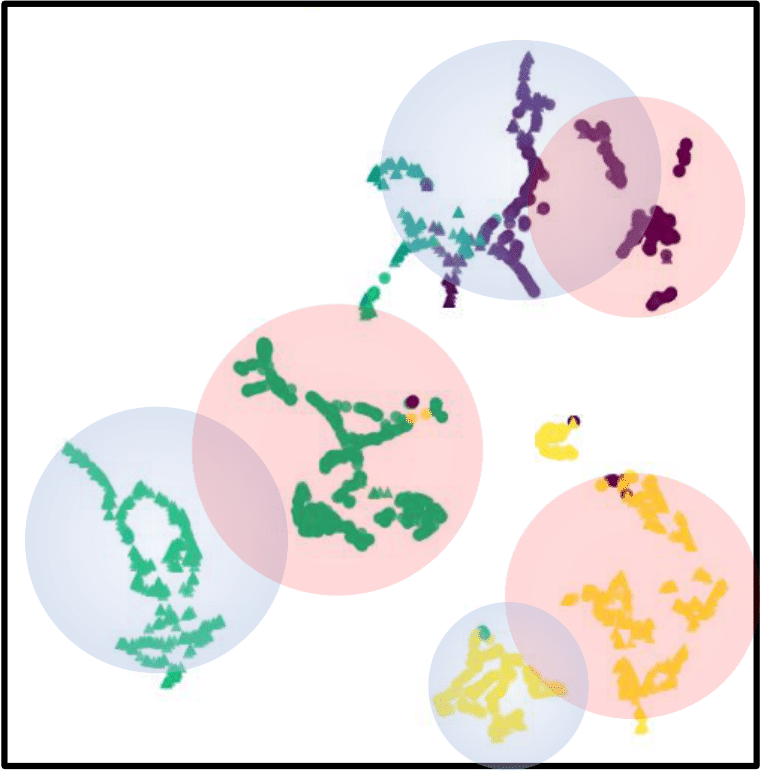}
        \captionsetup{font={small}}   
        \centering
        \caption{Baseline.}
     \end{subfigure}
     \begin{subfigure}{0.5\textwidth}
        \centering
        \includegraphics[width=1\linewidth]{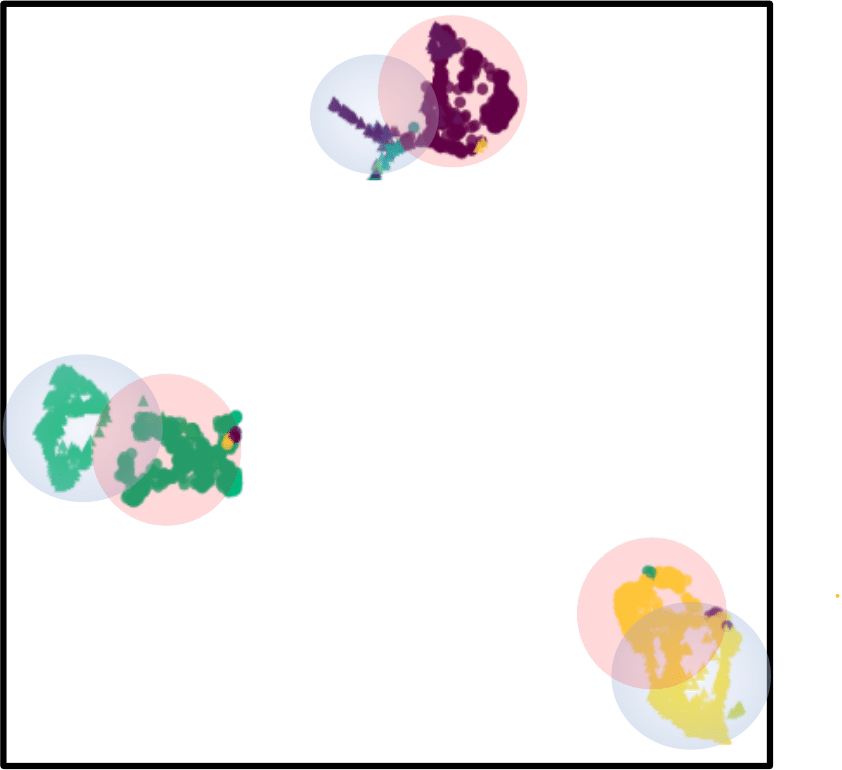}
        \captionsetup{font={small}}
        \centering
        \caption{Ours.}
     \end{subfigure}     
     \vspace{-2mm} 
     \captionsetup{font={small}}
     \caption{t-SNE visualization of a baseline~(a) and our model~(b). (Best viewed in color)}
     \label{fig:t-sne}
  \end{minipage} \hfill
  \begin{minipage}{0.45\linewidth}
    \captionsetup{font={small}}
    \captionof{table}{Quantitative results for variants of our model. We report mIoU scores for 19 and 16 classes on GTA5 $\rightarrow$ Cityscapes and SYNTHIA $\rightarrow$ Cityscapes, respectively.}
    \vspace{-1mm}
    \begin{adjustbox}{width=1\columnwidth,center} 
        \centering
        \large
        \begin{tabular}{ccccccc}
           \hline
           \multirow{2}{*}{$\mathcal{L}_\textrm{base}$  } & \multirow{2}{*}{$\mathcal{L}_{FC}$  } & \multirow{2}{*}{$\mathcal{L}_{BC}$} & \multirow{2}{*}{\begin{tabular}[c]{@{}c@{}}$+y_{\textrm{D}}$ \\ (w/o cal.) \end{tabular}} & \multirow{2}{*}{\begin{tabular}[c]{@{}c@{}}$+y_{\textrm{D}}$ \\ (w/ cal.)\end{tabular}} & \multicolumn{2}{c}{Source dataset}  \\ \cline{6-7} 
                                                        &                                     &                                     &                                                                          &                                                                                    & \multicolumn{1}{c}{GTA5$\text{       }$} & SYNTHIA \\ \hline
           \checkmark                                   &                                     &                                     &                                                                          &                                                                                    & \multicolumn{1}{c}{49.5} & 45.1    \\
           \checkmark                                   & \checkmark                          &                                     &                                                                          &                                                                                    & \multicolumn{1}{c}{51.2} & 48.8    \\
           \checkmark                                   & \checkmark                          & \checkmark                          &                                                                          &                                                                                    & \multicolumn{1}{c}{53.5} & 51.3    \\
           \checkmark                                   & \checkmark                          & \checkmark                          & \checkmark                                                               &                                                                                    & \multicolumn{1}{c}{55.3} & 53.5    \\
           \checkmark                                   & \checkmark                          & \checkmark                          &                                                                & \checkmark                                                                         & \multicolumn{1}{c}{57.1} & 55.6    \\ \hline
           \end{tabular}
    \end{adjustbox}
    \label{Tab:Ablation}
    
\end{minipage}
\vspace{-6mm}
\end{figure}

\vspace{-4mm}
\subsection{Results}    
\vspace{-2mm}
\subsubsection{Quantitative results.} We compare our method with the state-of-the-art methods on GTA5 $\rightarrow$ Cityscapes and SYNTHIA $\rightarrow$ Cityscapes in Tables~\ref{tab:gta} and~\ref{tab:synthia}, respectively. Note that all methods in the tables are based on the DeepLab-V2~\cite{chen2014semantic} architecture with ResNet-101, except for CAG-UDA~\cite{zhang2019category}. 
For a fair comparison, we report the results of ProDA~\cite{zhang2021prototypical} using the same network architecture as other methods, reproduced using an official source code. CBST~\cite{zou2018unsupervised} uses a self-training-based method to perform a class-aware alignment. This method is similar to our baseline, but it uses a limited number of pseudo labels, being outperformed by our approach on both benchmarks. 
PLCA~\cite{kang2020plca} uses a pixel-wise association method to align source and target domains in a pixel-level. This method, however, fails to obtain compact feature representations, and it is hence outperformed by our approach on both benchmarks. CorDA~\cite{wang2021domain} uses depth maps of source and target domains to transfer the knowledge of a source domain to a target one. Our method outperforms CorDA~\cite{wang2021domain} on both benchmarks even without using the depth information, indicating that our contrastive learning framework effectively transfers the knowledge across domains using pseudo labels. ProDA~\cite{zhang2021prototypical} focuses on removing false-positives of pseudo labels~\cite{zou2018unsupervised} and uses sparse labels. Different from ProDA~\cite{zhang2021prototypical}, we are interested in generating additional labels based upon the ones obtained by the approach of ~\cite{zou2018unsupervised}. That is, our method focuses on obtaining more true-positives and generating denser labels using pixel-prototype correspondences. Other than ProDA~\cite{zhang2021prototypical}, we additionally use the bi-directional contrastive loss to minimize intra-class variations and maximize inter-class variations of pixel-level features. We achieve mIoU gains of 0.6\% and 1.6\% for GTA5 $\rightarrow$ Cityscapes and SYNTHIA $\rightarrow$ Cityscapes, respectively, compared to ProDA~\cite{zhang2021prototypical}. The results imply that our method effectively learns domain-invariant and discriminative representations with denser pseudo labels, improving the mIoU performance of semantic segmentation. Additional comparisons with ProDA~\cite{zhang2021prototypical} are available in the supplement. We also report mIoU scores for the test split of Cityscapes, obtained from an official evaluation server, which has been ignored by most previous works. We use official source codes provided by the authors to obtain the results of state-of-the-art methods. We achieve non-trivial mIoU gains over CorDA~\cite{wang2021domain} and ProDA~\cite{zhang2021prototypical} for the test split of Cityscapes, demonstrating that ours can generalize better than them. Considering the performance gains of recent UDASS methods, the results are significant. For example, FDA~\cite{yang2020fda} achieves a mIoU gain of 0.3\% over CAG\_UDA~\cite{zhang2019category}, and TPLD~\cite{shin2020two} gets the gain of 0.7\% over FDA~\cite{yang2020fda}. CorDA~\cite{wang2021domain} and ProDA~\cite{zhang2021prototypical} provide large mIoU gains compared to other methods, but the improvements mainly come from exploiting additional depth maps and applying post-processing method, respectively.

\vspace{-5mm}
\subsubsection{Qualitative results.}
We show in Fig.~\ref{fig:segmentation} segmentation results on the GTA5 $\rightarrow$ Cityscapes task. Compared to the baseline model, our model provides more accurate segmentation results~(\eg, the bus in the first row, and the road and the rider in the second row). We show in Fig.~\ref{fig:t-sne} the t-SNE plot of feature representations of our model and the baseline. We visualize features of source and target images for each method by red and blue circles, respectively. The results show that our method successfully aligns the features for the same object category and separates them for different ones. That is, it minimizes intra-class variations, and maximizes inter-class variations, regardless of domains.

\vspace{-5mm} 
\subsubsection{Ablation study.} We present in Table~\ref{Tab:Ablation} an ablation analysis for each component of our framework on GTA5 $\rightarrow$ Cityscapes and SYNTHIA $\rightarrow$ Cityscapes. We show mIoU scores for variants of our model on the validation split of Cityscapes. As a baseline in the first row, we use static pseudo labels, obtained by the method of \cite{zou2018unsupervised}, to perform a class-aware alignment between source and target domains. We can see from the second row that FCL gives better mIoU scores, demonstrating the effectiveness of our approach to aligning prototypes and pixel-level features across domains. From the first and third rows, we can clearly see that jointly optimizing two contrastive terms is effective to UDASS. The fourth row demonstrates that leveraging additional dynamic pseudo labels provides better results than exploiting the static ones alone in terms of the mIoU score, even without the calibration~(w/o cal.). We can observe from the fifth row that the calibration~(w/ cal.) reduces domain discrepancies, and further improves the performance significantly.

\vspace{-5mm} 
\subsubsection{Comparison of pseudo labels.}
\begin{figure}[t]
  \begin{minipage}{0.6\linewidth}
    \begin{subfigure}{0.307\textwidth}
      \centering
      \includegraphics[width=1\linewidth]{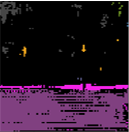}
      \captionsetup{font={small}}   
      \caption{w/o cal.}
   \end{subfigure} 
   \begin{subfigure}{0.313\textwidth}
      \centering
      \includegraphics[width=1\linewidth]{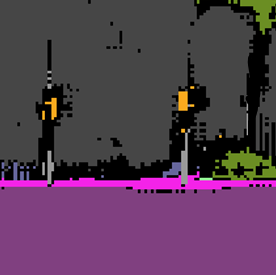}
      \captionsetup{font={small}}
      \caption{w/ cal.}
   \end{subfigure}
   \begin{subfigure}{0.328\textwidth}
      \centering
      \includegraphics[width=1\linewidth]{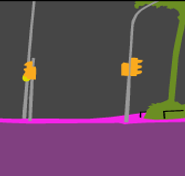}
      \captionsetup{font={small}}
      \caption{GT labels.}
   \end{subfigure}   
   \vspace{-3mm}
  \captionsetup{font={small}}
   \caption{Visualization of dynamic pseudo labels. (a-b) Pseudo labels obtained without and with calibrating prototypes of a source domain; (c) Target labels.}
   \label{fig:pseudolabel}
  \end{minipage} \hfill
  \begin{minipage}{0.37\linewidth}
    \captionsetup{font={small}}
    \captionof{table}{Quantitative results for various pseudo labels of a target domain. We report the densities of static, dynamic, and hybrid pseudo labels and corresponding label accuracies.}
    \vspace{-1mm}
    \begin{adjustbox}{width=1\columnwidth,center} 

        \centering
        \large
        \begin{tabular}{ C{2.8cm} C{2.1cm} C{2.5cm}}
          \hline
          \multicolumn{1}{c}{Pseudo labels} & Density(\%) & Accuracy(\%) \\ \hline 
          Static ~\cite{zou2018unsupervised}  & 20.1        & 98.5         \\
          Dyn. (w/o cal.)               & 22.2        & 98.6         \\
          Dyn. (w/ cal.)                & 34.3        & 98.6         \\
          Hybrid                             & 42.3        & 98.8         \\ \hline
          \end{tabular}
      \label{tab:pseudo label}
    \end{adjustbox}
\end{minipage}
\end{figure}

We measure the densities of various pseudo labels and corresponding label accuracies, and report the results in Table~\ref{tab:pseudo label}. We can see that the densities of dynamic pseudo labels are slightly higher than that of a static one, even without calibrating domain biases, while maintaining the label accuracies. Using pixel-prototype correspondences between target and source domains leads to obtaining denser labels than \cite{zou2018unsupervised}. The calibration process largely densifies dynamic pseudo labels. We can establish more correct correspondences between source and target domains by using the calibration process. The approach of~\cite{zou2018unsupervised} neglects the biases between source and target domains. Different from~\cite{zou2018unsupervised}, ours compensate for the class-wise domain biases and generate more accurate and denser labels than \cite{zou2018unsupervised}. Hybrid pseudo labels that combine static and dynamic ones provide the best result in terms of the label density and accuracy. When obtaining hybrid pseudo labels, we can reduce the number of incorrect static labels~\cite{zou2018unsupervised} by comparing them with dynamic ones. We show in Fig.~\ref{fig:pseudolabel} examples of dynamic pseudo labels obtained with and without the class-wise calibration. The results show that calibrating class-wise domain biases for source prototypes leads to establishing more correct pixel-prototype correspondences, providing denser and more accurate pseudo labels.

\begin{figure}[t]
  \centering
    \begin{subfigure}{0.42\textwidth}
      \centering
      \includegraphics[width=1\linewidth]{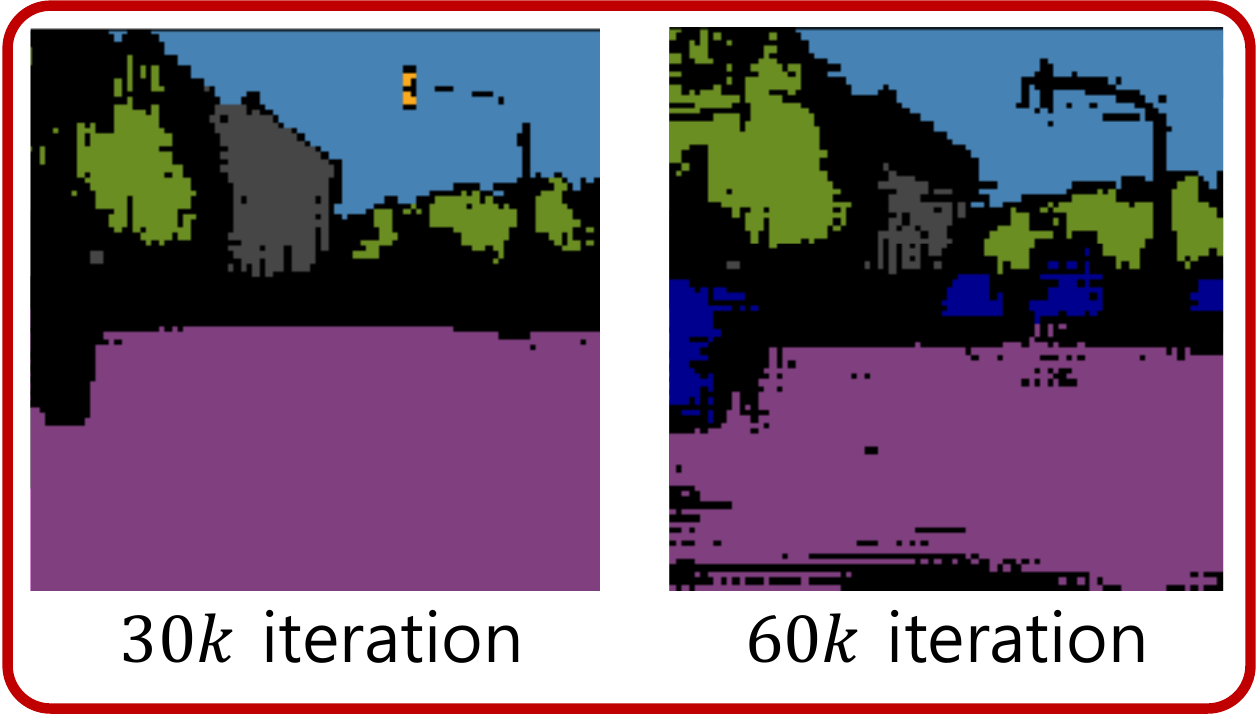}
      \captionsetup{font={small}}   
      \caption{Using $\rho_\textrm{S}$.}
   \end{subfigure} 
   \begin{subfigure}{0.42\textwidth}
      \centering
      \includegraphics[width=1\linewidth]{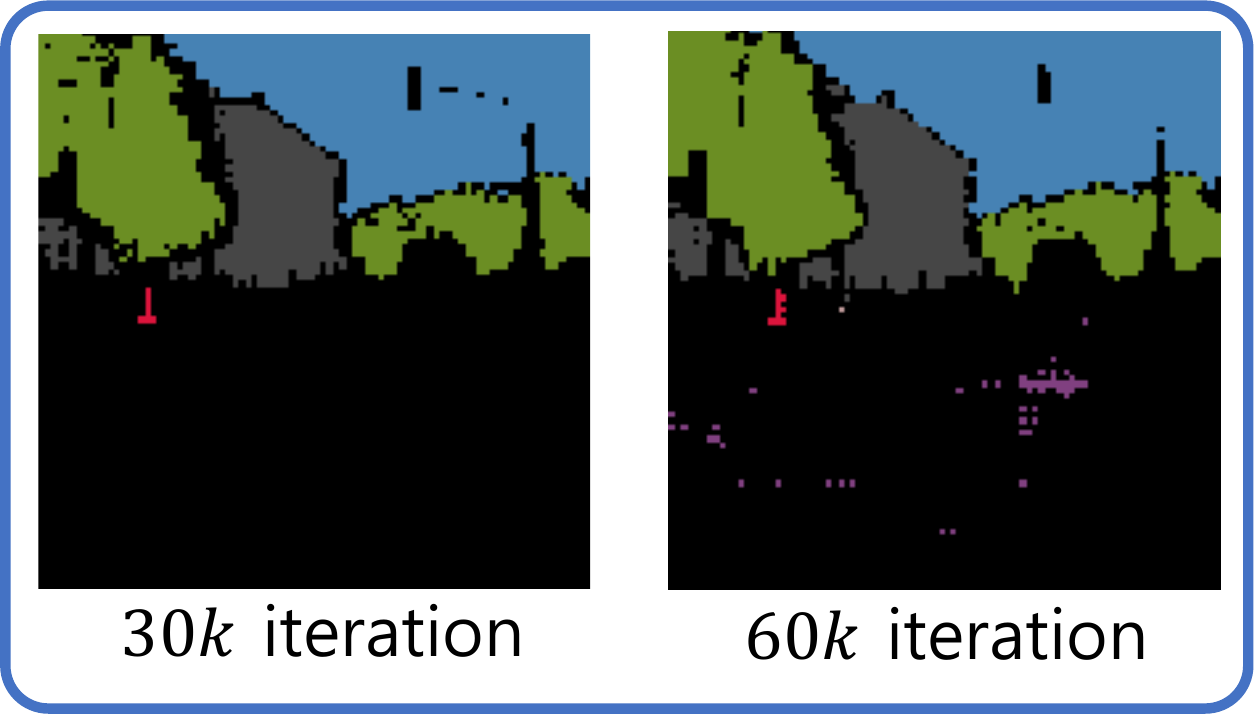}
      \captionsetup{font={small}}
      \caption{Using $\mu_\textrm{S}$.}
   \end{subfigure}
  \captionsetup{font={small}}
  \vspace{-2mm}
   \caption{Pseudo labels at 30$k$ and 60$k$ iterations using $\rho_\textrm{S}$ (a) and $\mu_\textrm{S}$ (b), respectively.}
   \vspace{-2mm}
   \label{fig:pseudolabel2}
\end{figure}

In Fig.~\ref{fig:pseudolabel2}, we compare generated pseudo labels using instance-wise prototypes~$\rho_\textrm{S}$ and momentum-based ones~$\mu_\textrm{S}$. We can see that using instance-wise prototypes~$\rho_\textrm{S}$ provides more diverse pseudo labels. They are more various than the other ones~$\mu_\textrm{S}$ slowly moving with momentum, and lead our model to establish diverse pixel-prototype correspondences.

\renewcommand{\arraystretch}{1.1}

\renewcommand{\arraystretch}{1.2}

\vspace{-3mm}
\section{Conclusion}
\vspace{-2mm}
We have introduced a novel contrastive learning framework for UDASS. Our key idea is to use cross-domain pixel-prototype correspondences to learn domain-invariant and discriminative representations. We have introduced a bi-directional contrastive loss  to align the features for the same object category and seperate them for different ones. We have also introduced an approach to generating pseudo labels dynamically in a nonparametric way using pixel-prototype correspondences, while compensating class-wise domain biases between source and target domains. Experimental results show the effectiveness of our framework, setting a new state of the art on standard benchmarks.

\vspace{-4mm}

\subsubsection{Acknowledgements.} This work was supported by Institute of Information \& communications Technology Planning \& Evaluation (IITP) grant funded by the Korea government (MSIT) (No.RS-2022-00143524, Development of Fundamental Technology and Integrated Solution for Next-Generation Automatic Artificial Intelligence System, and No.2022-0-00124, Development of Artificial Intelligence Technology for Self-Improving Competency-Aware Learning Capabilities), and the Yonsei Signature Research Cluster Program of 2022 (2022-22-0002).
\clearpage
%
%
\bibliographystyle{splncs04}
\bibliography{egbib}
\clearpage



\section*{Supplementary material}

# Bi-directional Contrastive Learning for Domain Adaptive Semantic Segmentation - Supplementary materials

Geon Lee, Chanho Eom, Wonkyung Lee, Hyekang Park, and Bumsub Ham*

Yonsei University

## 1 Training

**Hyperparameters.** Following [3], we set $\lambda_{seg}^{\mathrm{S}} = \lambda_{seg}^{\mathrm{T}} = 1$, and $\lambda_{ent}^{\mathrm{S}} = \lambda_{ent}^{\mathrm{T}} = 0.4$. To set the other hyperparameters, we divide the training set of Cityscapes [2] into two subsets. Specifically, we divide the images in the training set into two subsets of sizes 2380/595, and use them for training/cross-validation splits. We find all hyperparameters on the case of GTA5 [4] → Cityscapes [2]. For the hyperparameters $\lambda_{FC}$ and $\lambda_{BC}$, we perform a grid search over $\lambda_{FC}$, $\lambda_{BC}$ $\in \{0.1, 0.3, 0.5, 0.7\}$. For the momentum parameter $\lambda$ and the threshold value $\mathcal{T}$, we use a grid search over $\lambda \in \{0.99, 0.999, 0.9999\}$, and $\mathcal{T} \in \{0.5, 0.6, 0.7, 0.8, 0.9\}$, respectively. The results are shown in Fig. 1. We set $\lambda_{FC} = 0.5$, $\lambda_{BC} = 0.5$, $\lambda = 0.999$, and $\mathcal{T} = 0.7$. We fix hyperparameters, and train our model on all cases including GTA5 [4] → Cityscapes [2] and SYNTHIA [5] → Cityscapes [2].

**Weighted sampling.** Although a pair-wise training scheme in our framework is useful for aligning cross-domain features, it may lead to a class imbalance problem due to low co-occurrence rates of rare object classes, when source and target images are sampled randomly. To mitigate this issue, we use a weighted sampling strategy that samples more images from rare object categories in a source domain. Specifically, we calculate the occurrence ratios for all labels in the source domain before training, and then compute sampling probabilities $P$ for all labels by inverting the occurrence ratios as follows:

$$P(c) = \frac{1}{\sum_{c=1}^{C} p(c)^{-1}} p(c)^{-1},$$ (1)

where we denote by $P(c)$ and $p(c)$ a sampling probability and an occurrence ratio of the $c$-th class, respectively. We divide the number of images containing the $c$-th category by the total number of images in a source dataset to obtain $p(c)$. $C$ is the number of classes. For sampling a source-target pair in training time, we first sample a category from the sampling probability distribution $P$, and then pick a source image that contains the category in its ground-truth label.

---

* Corresponding author.

**Self-distillation.** Following [6, 7], we additionally apply a self-distillation technique to our final model. Specifically, we train the model using the method of [1], and use it as an initial model for the student model. We then apply a knowledge distillation, transferring the knowledge of our final model to the student which is pre-trained in a self-supervised manner. The loss for a distillation stage is as follows:

$$\mathcal{L}_{\text{KD}} = \mathcal{L}_{\text{base}} + \text{KL}(q_t || q_s), \tag{2}$$

where $q_t$, and $q_s$ denote the outputs of the teacher and student models, respectively. $\text{KL}(\cdot || \cdot)$ computes KL divergence of two predictions.

## 2   More experimental results

**Accuracy-density plots.** Static pseudo labels can be densified by lowering a threshold, but the label accuracy decreases accordingly. We provide an accuracy-density plot for static/dynamic pseudo labels in Fig. 2. We can see that dynamic pseudo labels are denser than static ones, while providing higher accuracies. We can establish more correct correspondences between source and target domains by using the calibration process. The approach of [8] neglects the biases between source and target domains. Different from [8], ours compensate for the class-wise domain biases and generate more accurate and denser labels than [8].

**Domain biases.** We estimate domain biases when obtaining dynamic pseudo labels. In Fig. 3, we visualize the magnitude of class-wise domain biases during training. The estimated domain biases gradually decrease during training, indicating that we successfully perform class-wise alignment between source and target domains.

**Quality of pseudo labels and mIoU performance.** We show the relationships between the quality of pseudo labels and mIoU performance in Table 1. By using static labels for contrastive learning, we obtain the mIoU performance of 53.5% for GTA5 → Cityscapes. And using dynamic ones for our learning leads to the mIoU improvements of 2%. By combining two labels, we generate hybrid labels, and using them leads to the mIoU performance of 57.1%. Using denser and more accurate labels contributes to the improvements of mIoU performance.

**Dynamic pseudo labels.** In Fig. 4, we visualize how dynamic pseudo labels change with iterations. We show the results at iteration $10k, 50k, 100k$, respectively. Dynamic pseudo labels are sparse and include vegetation, sky, and road classes at iteration $10k$. At iteration $50k$, our pseudo labels become denser and include additional classes of car and building. At iteration $100k$, our pseudo labels become more dense and accurate than labels at iteration $50k$, guiding better class-wise alignment.

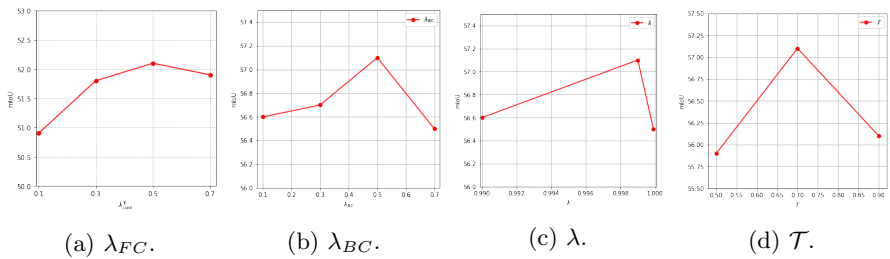

(a) $\lambda_{FC}$.     (b) $\lambda_{BC}$.     (c) $\lambda$.     (d) $\mathcal{T}$.

Fig. 1: (a-d) Quantitative comparisons for the hyperparameters, $\lambda_{FC}$, $\lambda_{BC}$, $\lambda$, and $\mathcal{T}$. We measure mIoU for all cases.

**Comparison with ProDA.** We compare the quality of pseudo labels for ours and ProDA in Table 2. ProDA [7] focuses on removing false-positives of pseudo labels [8] and uses sparse labels. Different from ProDA [7], we are interested in obtaining more true-positives and generating denser labels using pixel-prototype correspondences. By exploiting denser labels for the bi-directional contrastive losses, we achieve mIoU gains of 0.6% and 1.6% for GTA5 → Cityscapes and SYNTHIA → Cityscapes, respectively, compared to ProDA [7].

**Performances on a different backbone.** We show in Table 3 the results obtained using FCNs with VGG16 as a backbone network. We compare the segmentation performance of a baseline and our model on GTA5 → Cityscapes and SYNTHIA → Cityscapes. We show mIoU scores for both models on the validation split of Cityscapes. We can clearly see that our method boosts the performance of the baseline by significant margins.

**Memory and runtime for training and testing.** In Table 4, we report a total amount of memory usage and runtime for a baseline and our model at training and test times. We report the results for GTA5 → Cityscapes. Compared with the baseline, our model requires additional memory and time, but only at training time, which brings significant performance gains. For example, our method can boost the segmentation performance of the baseline by 7.6% for GTA5 → Cityscapes. Note that ours do not require extra memory and runtime at test time.

**Segmentation results.** We provide more visual results for GTA5 [4] → Cityscapes [2] in Fig. 5 and SYNTHIA [5] → Cityscapes [2] in Fig. 6. We compare the results of our model and those of the baseline. We can see that our model provides better segmentation outputs than the baseline for the classes of car, road, sidewalk, and rider. Our baseline model uses pseudo labels to perform the class-aware alignment. However, it does not consider intra-/inter-class variations of features. Our model learns discriminative features using the contrastive framework, and yields better segmentation results than the baseline.

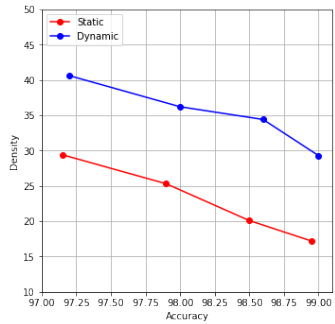

Fig. 2: An accuracy-density plot.

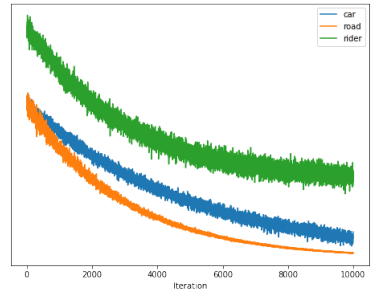

Fig. 3: Magnitude of domain biases for the classes of car, road, and rider.

| Pseudo labels | Density(%) | Accuracy(%) | mIoU(%) |
|---|---|---|---|
| Static [8] | 20.1 | 98.5 | 53.5 |
| Dyn. (w/o cal.) | 22.2 | 98.6 | 54.4 |
| Dyn. (w/ cal.) | 34.3 | 98.6 | 55.5 |
| Hybrid | 42.3 | 98.8 | 57.1 |

Table 1: Quantitative comparisons of pseudo labels (in terms of density and accuracy), and segmentation results (in terms of mIoU scores) using each label.

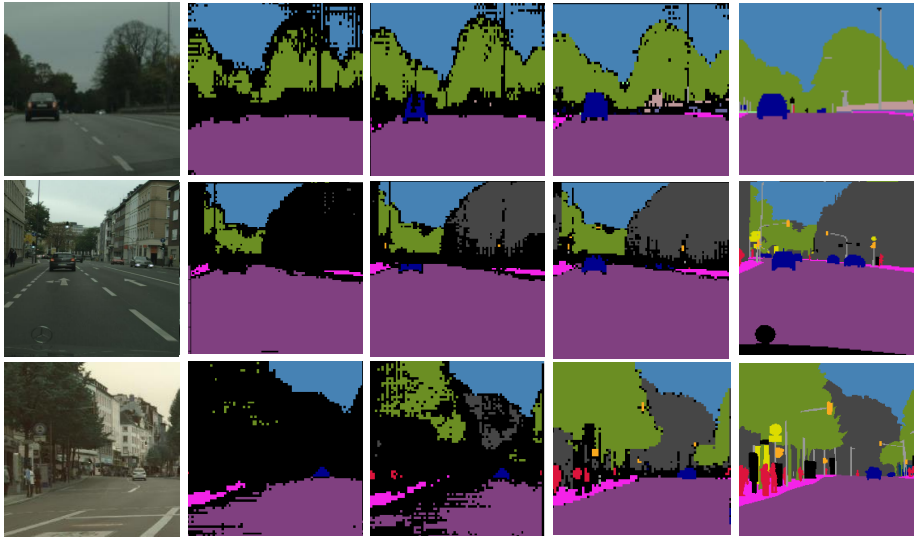

(a) Tgt. images.    (b) Iter 10$k$.    (c) Iter 50$k$.    (d) Iter 100$k$.    (e) GT labels.

Fig. 4: Visualizations of dynamic pseudo labels. (a) Target images. (b-d) Dynamic pseudo labels at iteration 10$k$, 50$k$, and 100$k$. (e) Ground-truth labels.

| Methods | Density(%) | Accuracy(%) |
|---------|-----------|-------------|
| ProDA [7] | 18.5 | 99.3 |
| Ours | 42.3 | 98.8 |

Table 2: Quantitative comparisons of pseudo labels (in terms of density and accuracy) for our method and ProDA [7].

| | GTA5 → Cityscapes | SYNTHIA → Cityscapes |
|---|---|---|
| Baseline | 41.1 | 43.2 |
| Ours | 48.5 | 52.4 |

Table 3: mIoU scores for a baseline and ours using the FCNs with VGG16 as a backbone network.

| | Memory usage (Train/Test) | Runtime (Train/Test) |
|---|---|---|
| Baseline | 24G/5G | 24 hours/13 minutes |
| Ours | 30G/5G | 32 hours/13 minutes |

Table 4: Quantitative comparisons of baseline and our models in terms of memory usage and runtime.

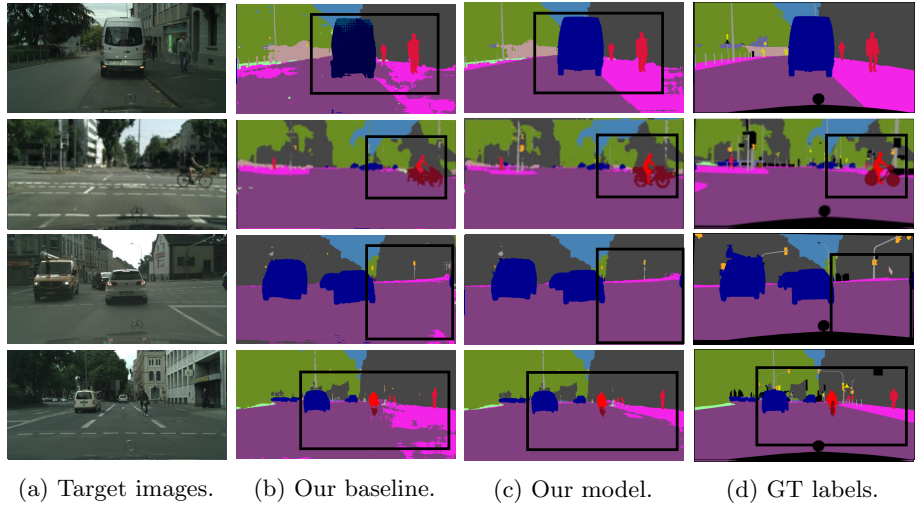

(a) Target images.     (b) Our baseline.     (c) Our model.     (d) GT labels.

Fig. 5: Qualitative comparisons on GTA5 → Cityscapes.

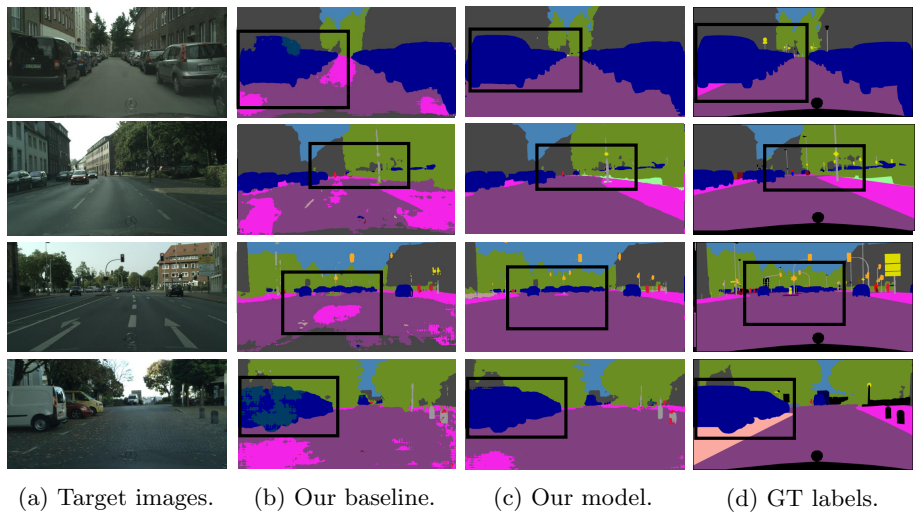

(a) Target images.     (b) Our baseline.     (c) Our model.     (d) GT labels.

Fig. 6: Qualitative comparisons on SYNTHIA → Cityscapes.



\end{document}